\documentclass[final,5p,times,twocolumn]{elsarticle}
\usepackage{etex}
\usepackage{ucs}
\usepackage{array}
\usepackage[utf8x]{inputenc}
\usepackage[T1]{fontenc}
\usepackage{lmodern}
\usepackage[normalem]{ulem}
\usepackage{booktabs}
\usepackage{amsmath,amssymb,amsthm}
\usepackage{bbm}
\usepackage{lineno,hyperref}\hypersetup{colorlinks=true,hypertexnames=false}
\modulolinenumbers[5]
\usepackage[osf,sc]{mathpazo}
\usepackage{graphicx}
\graphicspath{{./Fig/}}
\usepackage{ulem} 

\usepackage[usenames,dvipsnames]{xcolor}\definecolor{bg}{rgb}{0.95,0.95,0.95}
\usepackage[english]{babel}
\usepackage{listings}
\usepackage{multirow}
\usepackage{tabularx}

\usepackage{algorithm}
\usepackage{algorithmic}

\usepackage{afterpage}

\usepackage{pgfplotstable}
\usepgfplotslibrary{fillbetween}
\usetikzlibrary{positioning}
\usetikzlibrary{fit}
\pgfplotsset{compat=newest}
\usepackage{colortbl}

\begin{document}

\begin{frontmatter}

\title{Hierarchical Bayesian image analysis: \\from low-level modeling to robust supervised learning\tnoteref{fund}}

\author[n7]{Adrien Lagrange\corref{cor1}}
\ead{adrien.lagrange@enseeiht.fr}
\author[ensat]{Mathieu Fauvel}
\ead{mathieu.fauvel@ensat.fr}
\author[cnes]{St\'{e}phane May}
\ead{stephane.may@cnes.fr}
\author[n7,iuf]{Nicolas Dobigeon}
\ead{nicolas.dobigeon@enseeiht.fr}
\address[n7]{University of Toulouse, IRIT/INP-ENSEEIHT Toulouse, BP 7122, 31071 Toulouse Cedex 7, France}
\address[ensat]{INRA, DYNAFOR, BP 32607, Auzeville-Tolosane, 31326 Castanet Tolosan, France}
\address[cnes]{CNES, DCT/SI/AP, 18 Avenue Edouard Belin, 31400 Toulouse, France}
\address[iuf]{Institut Universitaire de France, France}
\cortext[cor1]{Corresponding author}
\tnotetext[fund]{Part of this work has been supported Centre National d'\'{E}tudes Spatiales (CNES), Occitanie Region and EU FP7 through the ERANETMED
JC-WATER program, MapInvPlnt Project ANR-15-NMED-0002-02.}

\begin{abstract}
Within a supervised classification framework, labeled data are used to learn classifier parameters. Prior to that, it is generally required to perform dimensionality reduction via feature extraction. These preprocessing steps have motivated numerous research works aiming at recovering latent variables in an unsupervised context. This paper proposes a unified framework to perform classification and low-level modeling jointly. The main objective is to use the estimated latent variables as features for classification and to incorporate simultaneously supervised information to help latent variable extraction. The proposed hierarchical Bayesian model is divided into three stages: a first low-level modeling stage to estimate latent variables, a second stage clustering these features into statistically homogeneous groups and a last classification stage exploiting the (possibly badly) labeled data. Performance of the model is assessed in the specific context of hyperspectral image interpretation, unifying two standard analysis techniques, namely unmixing and classification.
\end{abstract}


\end{frontmatter}


\section{Introduction}
\label{sec:intro}

  In the context of image interpretation, numerous methods have been developed to extract meaningful information. Among them, generative models have received a particular attention due to their strong theoretical background and the great convenience they offer in term of interpretation of the fitted models compared to some model-free methods such as deep neural networks. These methods are based on an explicit statistical modeling of the data which allows very task-specific model to be derived~\cite{Won2013book}, or either more general models to be implemented to solve generic tasks, such as Gaussian mixture model for classification~\cite{Kersten2014}. Task-specific and classification-like models are two different ways to reach an interpretable description of the data with respect to a particular applicative non-semantic issue. For instance, when analyzing images, task-specific models aim at recovering the latent (possibly physics-based) structures underlying each pixel-wise measurement \cite{Dobigeon2008} while classification provides a high-level information, reducing the pixel characterization to a unique label \cite{Fauvel2012}.

  Classification is probably one of the most common way to interpret data, whatever the application field of interest~\cite{Hastie2009}. This undeniable appeal has been motivated by the simplicity of the resulting output. This simplicity induces the appreciable possibility of benefiting from training data at a relatively low cost. Indeed, in general, experts can generally produce a ground-truth equivalent to the expected results of the classification for some amount of the data. This supervised approach allows a priori knowledge to be easily incorporated to improve the quality of the inferred classification model. Nevertheless, supervised methods are significantly influenced by the size of the training set, its representativeness and reliability~\cite{Foody2006}. Moreover, in some extent, modeling the pixel-wise data by a single descriptor may appear as somehow limited. It is the reason why the user-defined classes often refer to some rather vague semantic meaning with a possible large intra-class variability. To overcome these issues, while simultaneously facing with theoretical limitations of the expected classifier ability of generalization \cite{Jimenez1998}, an approach consists in preceding the training stage with feature extraction \cite{Sheikhpour2017}. These feature extraction techniques, whether parametric or nonparametric, have also the great advantage of simultaneously and significantly reducing the data volume to be handled as well as the dimension of the space in which the training should be subsequently conducted. Unfortunately, they are generally conducted in a separate manner before the classification task, i.e., without benefiting from any prior knowledge available as training data. Thus, a possible strategy is to consider a (possibly huge) set of features and selecting the relevant ones by appropriate optimization schemes \cite{Lagrange2017}.

  This observation illustrates the difficulty of incorporating ground-truthed information into a feature extraction step or, more generally, into a latent (i.e., unobserved) structure analysis. Due to the versatility of the data description, producing expert ground-truth with such degrees of accuracy and flexibility would be time-consuming and thus prohibitive. For example, for a research problem as important and well-documented as that of source separation, only very few and recent attempts have been made to incorporate supervised knowledge provided by an end-user \cite{Ozerov2011icassp}. Nonetheless, latent structure analysis may offer a relevant and meaningful interpretation of the data, since various conceptual yet structured knowledge to be inferred can be incorporated into the modeling. In particular, when dealing with measurements provided by a sensor, task-related biophysical considerations may guide the model derivation~\cite{Pereyra2012}. This is typically the case when spectral mixture analysis is conducted to interpret hyperspectral images whose pixel measurements are modeled as combinations of elementary spectra corresponding to physical elementary components~\cite{Bioucas-Dias2012}.

  The contribution of this paper lies in the derivation of a unified framework able to perform classification and latent structure modeling jointly. First,  this framework has the primary advantage of recovering consistent high and low level image descriptions, explicitly conducting hierarchical image analysis. Moreover, improvements in the results associated with both methods may be expected thanks to the complementarity of the two approaches. The use of ground-truthed training data is not limited to driving the high level analysis, i.e., the classification task. Indeed, it also makes it possible to inform the low level analysis, i.e., the latent structure modeling, which usually does not benefit well from such prior knowledge. On the other hand, the latent modeling inferred from each data as low level description can be used as features for classification. A direct and expected side effect is the explicit dimension reduction operated on the data before classification~\cite{Jimenez1998}. Finally, the proposed hierarchical framework allows the classification to be robust to corruption of the ground-truth. As mentioned previously, performance of supervised classification may be questioned by the reliability in the training dataset since it is generally built by human expert and thus probably corrupted by label errors resulting from ambiguity or human mistakes. For this reason, the problem of developing classification methods robust to label errors has been widely considered in the community~\cite{Bouveyron2009,Pelletier2017}. Pursuing this objective, the proposed framework also allows training data to be corrected if necessary.

  The interaction between the low and high level models is handled by the use of non-homogeneous Markov random fields (MRF)~\cite{Li2009}. MRFs are probabilistic models widely-used to describe spatial interactions. Thus, when used to derive a prior model within a Bayesian approach, they are particularly well-adapted to capture spatial dependencies between the latent structures underlying images~\cite{Zhang2001,Tarabalka2010}. For example, Chen \emph{et al.}~\cite{Chen2017} proposed to use MRFs to perform clustering. The proposed framework incorporates two instances of MRF, ensuring consistency between the low and high level modeling, consistency with external data available as prior knowledge and a more classical spatial regularization.

  The remaining of the article is organized as follows. Section~\ref{sec:prob} presents the hierarchical Bayesian model proposed as a unifying framework to conduct low-level and high-level image interpretation. A Markov chain Monte Carlo (MCMC) method is derived in Section~\ref{sec:mcmc} to sample according to the joint posterior distribution of the resulting model parameters. Then, a particular and illustrative instance of the proposed framework is presented in Section~\ref{sec:appli} where hyperspectral images are analyzed under the dual scope of unmixing and classification. Finally, Section~\ref{sec:ccl} concludes the paper and opens some research perspectives to this work.

\section{Bayesian model}
\label{sec:prob}

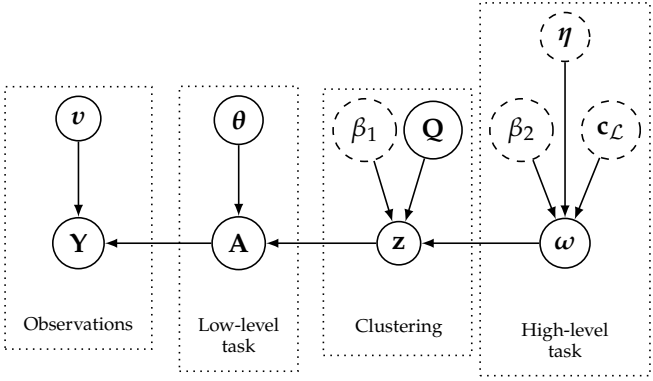
\begin{figure}[!ht]
    \centering
    \begin{tikzpicture}[auto, semithick, level 1/.style={sibling distance=0.9cm},edge from parent/.style={draw,latex-}]
      \node[circle,draw] (A) {$\mathbf{Y}$} [grow'=up]
        child {node[circle,draw,above=of A] (B) {$\boldsymbol{\upsilon}$}}
        ;

      \node[circle,draw,right=1.4cm of A] (E) {$\mathbf{A}$} [grow'=up]
        child {node[circle,draw,above=-0.2cm] (G) {$\boldsymbol{\boldsymbol{\theta}}$}};

      \node[circle,draw,right=1.45cm of E] (H) {$\mathbf{z}$} [grow'=up]
        child {node[circle,dashed,draw] (I) {$\beta_1$}}
        child {node[circle,draw] (J) {$\mathbf{Q}$}};

      \node[circle,draw,right=1.55cm of H] (K) {$\boldsymbol{\omega}$} [grow'=up]
        child {node[circle,dashed,draw,right=-0.1cm] (L) {$\beta_2$}}
        child {node[circle,dashed,draw,above=0.9cm] (M) {$\boldsymbol{\eta}$}}
        child {node[circle,dashed,draw,left=-0.1cm] (N) {$\mathbf{c}_\mathcal{L}$}};

      \draw[-latex] (E) -- (A);
      \draw[-latex] (H) -- (E);
      \draw[-latex] (K) -- (H);

      \node[below=0.5cm of A, font=\scriptsize] (L1) {Observations};
      \node[below=0.5cm of E, align=center, font=\scriptsize] (L2) {Low-level \\ task};
      \node[below=0.57cm of H, font=\scriptsize] (L3) {Clustering};
      \node[below=0.58cm of K, align=center, font=\scriptsize] (L4) {High-level \\ task};

      \node[draw,dotted,fit=(A) (B) (L1)] {};
      \node[draw,dotted,fit=(E) (G) (L2)] {};
      \node[draw,dotted,fit=(H) (I) (J) (L3)] {};
      \node[draw,dotted,fit=(K) (L) (M) (N) (L4)] {};

    \end{tikzpicture}
    \caption{Directed acyclic graph of the proposed hierarchical Bayesian model. (User-defined parameters appear in dotted circles).\label{fig:model}}
  \end{figure}

  In order to propose a unifying framework offering multi-level image analysis, a hierarchical Bayesian model is derived to relate the observations and the task-related parameters of interest. This model is mainly composed of three main levels. The first level, presented in Section~\ref{sec:low-task}, takes care of a low-level modeling achieving latent structure analysis. The second stage then assumes that data samples (e.g., resulting from measurements) can be divided into several statistically homogeneous clusters through their respective latent structures. To identify the cluster memberships, these samples are assigned discrete labels which are a priori described by a non-homogeneous Markov random field (MRF). This MRF combines two terms: the first one is related to the potential of a Potts-MRF to promote spatial regularity between neighboring pixels; the second term exploits labels from the higher level to promote coherence between cluster and classification labels. This clustering process is detailed in Section~\ref{sec:cluster}. Finally, the last stage of the model, explained in Section~\ref{sec:high-task}, allows high-level labels to be estimated, taking advantage of the availability of external knowledge as ground-truthed or expert-driven data, akin to a conventional supervised classification task. The whole model and its dependence is summarized by the directed acyclic graph in Figure~\ref{fig:model}.

\subsection{Low-level interpretation}
\label{sec:low-task}

  The low-level task aims at inferring $P$ $R$-dimensional latent variable vectors $\mathbf{a}_p$ ($\forall p \in \mathcal{P}  \triangleq \left\{1,\ldots,P\right\}$) appropriate for representing $P$ respective $d$-dimensional observation vectors $\mathbf{y}_p$ in a subspace of lower dimension than the original observation space, i.e., $R \leq d$. The task may also include the estimation of the function or additional parameters of the function relating the unobserved and observed variables. By denoting $\mathbf{Y} = \left[\mathbf{y}_1,\ldots,\mathbf{y}_P\right]$ and $\mathbf{A} = \left[\mathbf{a}_1,\ldots,\mathbf{a}_P\right]$  the $d \times P$- and $R \times P$- matrices gathering respectively the observation and latent variable vectors, this relation can be expressed through the general statistical formulation
  \begin{equation}
  \label{eq:model_lat}
    \mathbf{Y}| \mathbf{A},\boldsymbol{\upsilon} \sim  \psi\left(\mathbf{Y}; f_{\mathrm{lat}}\left(\mathbf{A}\right), \boldsymbol{\upsilon}\right),
  \end{equation}
  where $\psi(\cdot,\boldsymbol{\upsilon})$ stands for a statistical model, e.g., resulting from physical or approximation considerations, $f_{\mathrm{lat}}(\cdot)$ is a deterministic function used to define the latent structure and $\boldsymbol{\upsilon}$ are possible additional nuisance parameters. In most applicative contexts aimed by this work, the model $\psi(\cdot)$ and function $f_{\mathrm{lat}}(\cdot)$ are separable with respect to the measurements assumed to be conditionally independent, leading to the factorization
  \begin{equation}
  \label{eq:model_lat_fact}
    \mathbf{Y}| \mathbf{A},\boldsymbol{\upsilon} \sim  \prod_{p=1}^P \psi\left(\mathbf{y}_p; f_{\mathrm{lat}}\left(\mathbf{a}_p\right), \boldsymbol{\upsilon}\right).
  \end{equation}
It is worth noting that this statistical model will explicitly lead to the derivation of the particular form of the likelihood function involved in the Bayesian model.

The choice of the latent structure related to the function $f_{\mathrm{lat}}(\cdot)$ is application-dependent and can be directly chosen by the end-user. A conventional choice consists in considering a linear expansion of the observed data $\mathbf{y}_p$ over an orthogonal basis spanning a space whose dimension is lower than the original one. This orthogonal space can be a priori fixed or even learnt from the dataset itself, e.g., leveraging on popular nonparametric methods such as principal component analysis (PCA)~\cite{Fauvel2006}. In such case, the model \eqref{eq:model_lat} should be interpreted as a probabilistic counterpart of PCA \cite{Tipping1999} and the latent variables $\mathbf{a}_p$ would correspond to factor loadings. Similar linear latent factors and low-rank models have been widely advocated to address source separation problems, such as nonnegative matrix factorization~\cite{Fevotte2009}. As a typical illustration, by assuming an additive white and centered Gaussian statistical model $\psi(\cdot)$ and a linear latent function $f_{\mathrm{lat}}(\cdot)$, the generic model \eqref{eq:model_lat_fact} can be particularly instanced as
\begin{equation}
\label{eq:model_lat_gauss}
 \mathbf{Y}| \mathbf{A},s^2 \sim  \prod_{p=1}^P \mathcal{N}\left(\mathbf{M}\mathbf{a}_p,s^2 \mathbf{I}_{d}\right)
\end{equation}
where $\mathbf{I}_{d}$ is the $d\times d$ identity matrix, $\mathbf{M}$ is a matrix spanning the signal subspace and $s^2$ is the variance of the Gaussian error, considered as a nuisance parameter. Besides this popular class of Gaussian models, this formulation allows other noise statistics to be handled within a linear factor modeling, as required when the approximation should be envisaged beyond a conventional Euclidean discrepancy measure \cite{Fevotte2011}, provided that
$$\mathbb{E}\left[\mathbf{Y}|\mathbf{A}\right] = f_{\mathrm{lat}}(\mathbf{A}).$$

From a different perspective, the generic formulation of the statistical latent structure \eqref{eq:model_lat_fact} can also result from a thorough analysis of more complex physical processes underlying observed measurements, resulting in specific yet richer physics-based latent models~\cite{Pereyra2012,Albughdadi2014}. For sake of generality, this latent structure will not be specified in the rest of this manuscript, except in Section \ref{sec:appli} where the linear Gaussian model \eqref{eq:model_lat_gauss} will be more deeply investigated as an illustration in a particular applicative context.

\subsection{Clustering}
\label{sec:cluster}
  To regularize the latent structure analysis, the model is complemented by a clustering step as a higher level of the Bayesian hierarchy. Besides, another objective of this clustering stage is also to act as a bridge between the low- and high-level data interpretations, namely latent structure analysis and classification. The clustering is performed under the assumption that the latent variables are statistically homogeneous and allocated in several clusters, i.e., identities belonging to a same cluster are supposed to be distributed according to the same distribution. To identify the membership, each observation is assigned a cluster label $z_p\in\mathcal{K} \triangleq \left\{1,\ldots,K\right\}$ where $K$ is the number of clusters. Formally, the unknown latent vector is thus described by the following prior
  \begin{equation}
  \label{eq:prior_lat}
    \mathbf{a}_p|z_p=k,\boldsymbol{\theta}_k \sim \Phi(\mathbf{a}_p;\boldsymbol{\theta}_k),
  \end{equation}
  where $\Phi$ is a given statistical model depending on the addressed problem and governed by the parameter vector $\boldsymbol{\theta}_k$ characterizing each cluster. As an example, considering this prior distribution as Gaussian, i.e., $\Phi(\boldsymbol{\theta}_k) = \mathcal{N}(\boldsymbol{\psi}_k,\boldsymbol{\Sigma}_k)$ with $\boldsymbol{\theta}_k = \left\{\boldsymbol{\psi}_k,\boldsymbol{\Sigma}_k\right\}$, would lead to a conventional Gaussian mixture model (GMM) for the latent structure, as in~\cite{Eches2011} (see Section~\ref{sec:appli}).\\

  One particularity of the proposed model lies in the prior on the cluster labels $\mathbf{z}=\left[{z}_1,\ldots,{z}_P\right]$. A non-homogeneous Markov Random Field (MRF) is used as a prior model to promote two distinct behaviors through the use of two potentials. The first one is a local and non-homogeneous potential parametrized by a $K$-by-$J$ matrix $\mathbf{Q}$. It promotes consistent relationships between the cluster labels $\mathbf{z}$ and some classification labels $\boldsymbol{\omega}=\left[{\omega}_1,\ldots,{\omega}_P\right]$ where $\omega_p\in\mathcal{J} \triangleq \left\{1,\ldots,J\right\}$ and $J$ is the number of classes. These classification labels associated with high-level interpretation will be more precisely investigated in the third stage of the hierarchy in Section \ref{sec:high-task}. Pursuing the objective of analyzing images, the second potential is associated with a Potts-MRF~\cite{Wu1982} of granularity parameter $\beta_1$ to promote a piecewise consistent spatial regularity of the cluster labels. The prior probability of $\mathbf{z}$ is thus defined as
  \begin{alignat}{2}
    \mathbb{P}&[\mathbf{z}|\boldsymbol{\omega},\mathbf{Q}] =&& \nonumber\\
    & \frac{1}{C(\boldsymbol{\omega},\mathbf{Q})} \exp&&\left( \sum_{p \in \mathcal{P}} V_{1}(z_p,\omega_p,q_{z_p,\omega_p}) \right. \nonumber\\
    &&& \left. + \sum_{p \in \mathcal{P}} \sum_{p^\prime \in \mathcal{V}(p)} V_2(z_p,z_{p^\prime }) \right)
  \end{alignat}
  where $\mathcal{V}(p)$ stands for the neighborhood of $p$, $q_{k,j}$ is the $k$-th element of the $j$-th column of $\mathbf{Q}$. The two terms $V_1(\cdot)$ and $V_2(\cdot)$ are the classification-informed and Potts-Markov potentials, respectively, defined by
  \begin{align*}
    &V_1(k,j,q_{k,j}) = \log(q_{k,j}) \\
    &V_2(k,k^\prime) = \beta_1 \delta(k,k^\prime)
  \end{align*}
  where $\delta(\cdot,\cdot)$ is the Kronecker function. Finally, $C(\boldsymbol{\omega},\mathbf{Q})$ stands for the normalizing constant (i.e., partition function) depending of $\boldsymbol{\omega}$ and $\mathbf{Q}$ and computed over all the possible $\mathbf{z}$ fields~\cite{Li2009}
  \begin{alignat}{2}
    C(\boldsymbol{\omega},&\mathbf{Q}) = \sum_{\mathbf{z} \in \mathcal{K}^P} \exp&& \left( \sum_{p \in \mathcal{P}} V_1(z_p,\omega_p,q_{z_p,\omega_p}) \right. \nonumber\\
    &&& \left.+ \sum_{p \in \mathcal{P}} \sum_{p^\prime \in \mathcal{V}(p)} V_2(z_p,z_{p^\prime}) \right) \nonumber\\
    &= \sum_{\mathbf{z} \in \mathcal{K}^P} \prod_{p \in \mathcal{P}} q_{z_p,\omega_p}&& \exp \left(\beta_1 \sum_{p^\prime \in \mathcal{V}(p)} \delta(z_p,z_{p^\prime})\right)
  \end{alignat}

  The equivalence between Gibbs random fields and MRF stated by the Hammersley-Clifford theorem~\cite{Li2009} provides the prior probability of a particular cluster label conditionally upon its neighbors
  \begin{align}
    \mathrm{P}&[z_p = k|\mathbf{z}_{\mathcal{V}(p)},\omega_p=j,q_{k,\omega_p}] \propto \nonumber\\
    & \exp \left( V_1(k,j,q_{k,j}) + \sum_{p^\prime \in \mathcal{V}(p)} V_2(k,z_{p^\prime}) \right)
  \end{align}
 where the symbol $\propto$ stands for ``proportional to''.

  The elements $q_{k,j}$ of the matrix $\mathbf{Q}$ introduced in the latter MRF account for the connection between cluster $k$ and class $j$, revealing a hidden interaction between clustering and classification. A high value of $q_{k,j}$ tends to promote the association to the cluster $k$ when the sample belongs to the class $j$. This interaction encoded through these matrix coefficients is unknown and thus motivates the estimation of the matrix $\mathbf{Q}$. To reach an interpretation of the matrix coefficients in terms of probabilities of inter-dependency, a Dirichlet distribution is elected as prior for each column $\mathbf{q}_j = \left[q_{1,j},\ldots,q_{K,j}\right]^T$ of $\mathbf{Q}=\left[\mathbf{q}_1,\ldots,\mathbf{q}_J\right]$ which are assumed to be independent, i.e.,
  \begin{equation}
  \label{eq:prior_Q}
    \mathbf{q}_{j} \sim \text{Dir}(\zeta_1,\ldots,\zeta_K).
  \end{equation}
  The nonnegativity and sum-to-one constraints imposed to the coefficients defining each column of $\mathbf{Q}$ allows them to be interpreted as probability vectors. The choice of such a prior is furthermore motivated by the properties of the resulting conditional posterior distribution of $\mathbf{q}_j$, as demonstrated later in Section~\ref{sec:mcmc}. In the present work, the hyperparameters $\zeta_1,\ldots,\zeta_K$ are all chosen equal to $1$, resulting in a uniform prior over the corresponding simplex defined by the probability constraints. Obviously, when additional prior knowledge on the interaction between clustering and classification is available, these hyperparameters can be adjusted accordingly.

\subsection{High-level interpretation}
\label{sec:high-task}

  The last stage of the hierarchical model defines a classification rule. At this stage, a unique discrete class label should be attributed to each sample. This task can be seen as high-level in the sense that the definition of the classes can be motivated by their semantic meaning. Classes can be specified by the end-user and thus a class may gather samples with significantly dissimilar observation vectors and even dissimilar latent features. The clustering stage introduced earlier also allows a mixture model to be derived for this classification task. Indeed, a class tends to be the union of several clusters identified at the clustering stage, providing a hierarchical description of the dataset.\\

  In this paper, the conventional and well-admitted setup of a supervised classification is considered. This setup means that a partial ground-truthed dataset $\mathbf{c}_\mathcal{L}$ is available for a (e.g., small) subset of samples. In what follows, $\mathcal{L} \subset \mathcal{P}$ denotes the subset of observation indexes for which this ground-truth is available. This ground-truth provides the expected classification labels for observations indexed by $\mathcal{L}$. Conversely, the index set of unlabeled samples for which this ground-truth is not available is noted $\mathcal{U} \subset \mathcal{P}$, with $\mathcal{P} = \mathcal{U} + \mathcal{L}$ and $\mathcal{U} \cap \mathcal{L} = \emptyset$. Moreover, the proposed model assumes that this ground-truth may be corrupted by class labeling errors. As a consequence, to provide a classification robust to these possible errors, all the classification labels of the dataset will be estimated, even those associated with the observations indexed by $\mathcal{L}$. At the end of the classification process, the labels estimated for observations indexed by $\mathcal{L}$ will not be necessarily equal to the labels $\mathbf{c}_\mathcal{L}$ provided by the expert or an other external knowledge.\\

  Similarly to the prior model advocated for $\mathbf{z}$ (see Section \ref{sec:cluster}), the prior model for the classification labels $\boldsymbol{\omega}$ is a non-homogeneous MRF composed of two potentials. Again, a Potts-MRF potential with a granularity parameter $\beta_2$ is used to promote spatial coherence of the classification labels. The other potential is non-homogeneous and exploits the supervised information available under the form of the ground-truth map $\mathbf{c}_\mathcal{L}$. In particular, it attends to ensure consistency between the estimated and ground-truthed labels for the samples indexed by $\mathcal{L}$. Moreover, for the classification labels associated with the indexes in $\mathcal{U}$ (i.e., for which no ground-truth is available), the prior probability to belong to a given class is set as the proportion of this class observed in $\mathbf{c}_\mathcal{L}$. This setting assumes that the expert map is representative of the whole scene to be analyzed in term of label proportions. If this assumption is not verified, the proposed modeling can be easily adjusted accordingly. Mathematically, this formal description can be summarized by the following conditional prior probability for a given classification label $\omega_p$
  \begin{align}
    \mathbb{P}[\omega_p=j&|\boldsymbol{\omega}_{\mathcal{V}(p)},c_p,\eta_p] \propto \nonumber \\
    &\exp \left( W_1(j,c_p,\eta_p) + \sum_{p^\prime \in \mathcal{V}(p)} W_2(j,\omega_{p^\prime}) \right).
  \end{align}
As explained above, the potential $W_2(\cdot,\cdot)$ ensures the spatial coherence of the classification labels, i.e.,
\begin{equation*}
    W_2(j,j^\prime) = \beta_2 \delta(j,j^\prime).
  \end{equation*}
More importantly, the potential $ W_1 (j, c_p, \eta_p)$ defined by
  \begin{align*}
    W_1&(j,c_p,\eta_p) =  \\
    &\begin{cases}
      \begin{cases}
        \log(\eta_p), & \text{when}\ j = c_p \\
        \log(\frac{1-\eta_p}{J-1}), & \text{otherwise}
      \end{cases}, & \text{when $p\in \mathcal{L}$}\\
       -\log(\pi_j), & \text{when $p\in \mathcal{U}$}
    \end{cases}
  \end{align*}
encodes the coherence between estimated and ground-truthed labels when available (i.e., when $p\in \mathcal{L}$)
or, conversely for non-ground-truthed labels  (i.e., when $p\in \mathcal{U}$), the prior probability of assigning a given label through the proportion $\pi_j$ of samples of class $j$ in $\mathbf{c}_\mathcal{L}$. The hyperparameter $\eta_p\in(0,1)$ stands for the confidence given in $c_p$, i.e., the ground-truth label of pixel $p$. In the case where the confidence is total, the parameter tends to $1$ and it leads to $\omega_p=c_p$ in a deterministic manner. However, in a more realistic applicative context, ground-truth is generally provided by human experts and may contain errors due for example to ambiguities or simple mistakes. It is possible with the proposed model to set for example a $90\%$ level of confidence which allows to re-estimate the class label of the labeled set $\mathcal{L}$ and thus to correct the provided ground-truth. By this mean, the robustness of the classification to label errors is improved.

\section{Gibbs sampler}
\label{sec:mcmc}

To infer the parameters of the hierarchical Bayesian model introduced in the previous section, an MCMC algorithm is derived to generate samples according to the joint posterior distribution of interest which can be computed according to the following hierarchical structure
\begin{align*}
&p\left(\mathbf{A}, \boldsymbol{\Theta}, \mathbf{z}, \mathbf{Q}, \boldsymbol{\omega}|\mathbf{Y}\right)\propto p(\mathbf{Y}|\mathbf{A})p(\mathbf{A}|\mathbf{z}, \boldsymbol{\theta})p(\mathbf{z}|\mathbf{Q},\boldsymbol{\omega})p(\boldsymbol{\omega})
\end{align*}
with $\boldsymbol{\Theta} \triangleq \left\{\boldsymbol{\theta}_1,\ldots,\boldsymbol{\theta}_K\right\}$. Note that, for conciseness, the nuisance parameters $\boldsymbol{\upsilon}$ have been implicitly marginalized out in the hierarchical structure. If this marginalization is not straightforward, these nuisance parameters can be also explicitly included within the model to be jointly estimated.

The Bayesian estimators of the parameters of interest can then be approximated using these samples. The minimum mean square error (MMSE) estimators of the parameters $\mathbf{A}$, $\boldsymbol{\Theta}$ and $\mathbf{Q}$ can be approximated through empirical averages
\begin{equation}
  \hat{\mathbf{x}}_{\mathrm{MMSE}} = \mathbb{E}[\mathbf{x}|\mathbf{Y},] \approx \frac{1}{N_{\mathrm{MC}}} \sum_{t=1}^{N_{\mathrm{MC}}} \mathbf{x}^{(t)}
\end{equation}
where $\cdot^{(t)}$ denotes the $t$th samples and $N_{\mathrm{MC}}$ is the number of iterations after the burn-in period. Conversely, the maximum a posteriori estimators of the cluster and class labels, $\mathbf{z}$ and $\boldsymbol{\omega}$, respectively, can be approximated as
\begin{equation}
   \hat{\mathbf{x}}_{\mathrm{MAP}} = \operatornamewithlimits{argmax}_{\mathbf{x}} p(\mathbf{x}|\mathbf{Y}) \approx \operatornamewithlimits{argmax}_{\mathbf{x}^{(t)}} p(\mathbf{x}^{(t)}|\mathbf{Y})
\end{equation}
which basically amounts at retaining the most frequently generated label for these specific discrete parameters \cite{Kail2012}.

To carry out such a sampling strategy, the conditional posterior distributions of the various parameters need to be derived. More importantly, the ability of drawing according to these distributions is required. These posterior distributions are detailed in what follows.

\subsection{Latent parameters}
Given the likelihood function resulting from the statistical model \eqref{eq:model_lat_fact} and the prior distribution in \eqref{eq:prior_lat}, the conditional posterior distribution of a latent vectors can be expressed as follows
  \begin{align}
  p(\mathbf{a}_p&|\mathbf{y}_p, \boldsymbol{\upsilon}, z_p=k, \boldsymbol{\theta}_k) \propto p(y_p|\mathbf{a}_p,\boldsymbol{\upsilon}) p(\mathbf{a}_p|z_p=k,\boldsymbol{\theta}_k) \nonumber\\
  &\propto \psi\left(\mathbf{y}_p; f_{\mathrm{lat}}\left(\mathbf{a}_p\right), \boldsymbol{\upsilon}\right) \Phi(\mathbf{a}_p;\boldsymbol{\theta}_k).
  \end{align}


\subsection{Cluster labels}
 The cluster label $z_p$ being a discrete random variable, it is possible to sample the variable by computing the conditional probability for all possible values of $z_p$ in $\mathcal{K}$
  \begin{align}
  \mathbb{P}&(z_p=k|\boldsymbol{\theta}_k,\omega_p=j,q_{k,j})  \nonumber\\
    &\propto p(\mathbf{a}_p|z_p=k,\boldsymbol{\theta}_k) \mathbb{P}(z_p=k|\mathbf{z}_{\mathcal{V}(p)},\omega_p=j,q_{k,j}) \nonumber\\
    &\propto \Phi(\mathbf{a}_p;\boldsymbol{\theta}_k) q_{k,j} \exp \left( \beta_1 \sum_{p^\prime \in \mathcal{V}(p)} \delta(k,z_p^\prime) \right) \label{eq:sampling_z}.
  \end{align}

\subsection{Interaction matrix}
  The conditional distribution of each column $\mathbf{q}_j$ ($j \in \mathcal{J}$) of the interaction parameter matrix $\mathbf{Q}$ can be written
  \begin{align}
  p(\mathbf{q}_j|\mathbf{z},\mathbf{Q}_{\backslash j}, \boldsymbol{\omega}) &\propto p(\mathbf{q}_j) \mathbb{P}(\mathbf{z}|\mathbf{Q},\boldsymbol{\omega}) \nonumber \\
  &\propto \frac{ \prod_{k=1}^{K} q_{k,j}^{n_{k,j}} }{C(\boldsymbol{\omega},\mathbf{Q})} \mathbbm{1}_\mathbb{S}(\mathbf{q}_j) \label{eq:q_condlaw_full}.
  \end{align}
  where $\mathbf{Q}_{\backslash j}$ denotes the matrix $\mathbf{Q}$ whose $j$th column has been removed, $n_{k,j}=\#\{p|z_p=k,\omega_p=j\}$ is the number of observations whose cluster and class labels are respectively $k$ and $j$, and $\mathbbm{1}_\mathbb{S}(\cdot)$ is the indicator function of the probability simplex which ensures that $\mathbf{q}_j \in \mathbb{S}$ implies $\forall k \in \mathcal{K}$, $q_{k,j} \geq 0$ and $\sum_{k=1}^{K} q_{k,j} = 1$.

  Sampling according to this conditional distribution would require to compute the partition function $C(\boldsymbol{\omega},\mathbf{Q})$, which is not straightforward. The partition function is indeed a sum over all possible configurations of the MRF $\mathbf{z}$. One strategy would consist in precomputing this partition function on an appropriate grid, as in \cite{Risser2010}. As alternatives, one could use to likelihood-free Metropolis Hastings algorithm \cite{Pereyra2013ip}, auxiliary variables \cite{Moller2006} or pseudo-likelihood estimators \cite{Besag1975}. However, all these strategies remain of high computational cost, which precludes their practical use for most applicative scenarii encountered in real-world image analysis.

  Besides, when $\beta_1 = 0$, this partition function reduces to $C(\boldsymbol{\omega},\mathbf{Q}) = 1$. In other words, the partition function is constant when the spatial regularization induced by $V_2(\cdot)$ is not taken into account. In such case, the conditional posterior distribution for $\mathbf{q}_j$ is the following Dirichlet distribution
  \begin{align}
  \label{eq:q_condlaw}
    \mathbf{q}_{j}|\mathbf{z},\boldsymbol{\omega} \sim \text{Dir}(n_{1,j}+1,\ldots,n_{K,j}+1),
  \end{align}
  which is easy to sample from. Interestingly, the expected value of $q_{k,j}$ is then
  $$\mathrm{E}\left[q_{k,j}|\mathbf{z},\boldsymbol{\omega} \right] = \frac{n_{k,j}+1}{\sum_{i=1}^{K} n_{i,k} + K}$$
   which is a biased empirical estimator of $\mathrm{P}\left[z_p=k|\omega_p=j\right]$. This latter result motivates the use of a Dirichlet distribution as a prior for $\mathbf{q}_j$. Thus, it is worth noting that $\mathbf{Q}$ can be interpreted as a byproduct of the proposed model which describes the intrinsic dataset structure. It allows the practitioner not only to get an overview of the distribution of the samples of a given class in the various clusters but also to possibly identify the origin of confusions between several classes. Again, this clustering step allows disparity in the semantic classes to be mitigated. Intraclass variability results in the emerging of several clusters which are subsequently agglomerated during the classification stage.

  In practice, during the burn-in period of the proposed Gibbs sampler, to avoid highly intensive computations, the cluster labels are sampled according to \eqref{eq:sampling_z} with $\beta_1>0$ while the columns of the interaction matrix are sampled according to \eqref{eq:q_condlaw}. In other words, during this burn-in period, a certain spatial regularization with $\beta_1>0$ is imposed to the cluster labels and the interaction matrix is sampled according to an approximation of its conditional posterior distribution\footnote{This strategy can also be interpreted as choosing $C(\boldsymbol{\omega},\mathbf{Q}) \times \text{Dir}(\mathbf{1})$ instead of the Dirichlet distribution \eqref{eq:prior_Q} as prior for $\mathbf{q}_j$.}. After this burn-in period, the granularity parameter $\beta_1$ is set to $0$, which results in removing the spatial regularization between the cluster labels. Thus, once convergence has been reached, the conditional posterior distribution \eqref{eq:q_condlaw} reduces to \eqref{eq:q_condlaw_full} and the iteraction matrix is properly sampled according to its exact conditional posterior distribution.

%

\subsection{Classification labels}
  Similarly to the cluster labels, the classification labels $\boldsymbol{\omega}$ are sampled by evaluating their conditional probabilities computed for all the possible labels. However, two cases need to be considered while sampling the classification label ${\omega}_p$, depending on the availability of ground-truth label for the corresponding $p$th pixel. More precisely, when $p \in \mathcal{U}$, i.e., when the $p$th pixel is not accompanied by a corresponding ground-truth, the conditional probabilities are written
  \begin{align}
    &\mathbb{P}[\omega_p=j|\mathbf{z}, \boldsymbol{\omega}_{\backslash p}, \mathbf{q}_j,c_p,\eta_p] \nonumber\\
    &\propto \mathbb{P}[z_p|\omega_p=j,\mathbf{q}_j,\mathbf{z}_{\nu(p)}] \mathbb{P}[\omega_p=j|\boldsymbol{\omega}_{\mathcal{V}(p)},c_p,\eta_p] \nonumber\\
    &\propto \frac{q_{z_p,j} \pi_j \exp \left( \beta_2 \sum_{p^\prime \in \nu(p)} \delta(j,\omega_{p^\prime}) \right)}{ \sum\limits_{k^\prime=1}^{K} q_{k^\prime,j} \exp \left( \beta_1 \sum\limits_{p^\prime \in \nu(p)} \delta(k^\prime,z_{p^\prime}) \right)},
  \end{align}
  where $\boldsymbol{\omega}_{\backslash p}$ denotes the classification label vector $\boldsymbol{\omega}$ whose $p$th element has been removed.
  Conversely, when $p \in \mathcal{L}$, i.e., when the $p$th pixel is assigned a ground-truth label $c_p$, the conditional posterior probability reads
  \begin{align}
    &\mathbb{P}[\omega_p=j|\mathbf{z}, \boldsymbol{\omega}_{\backslash p}, \mathbf{q}_j,c_p,\eta_p] \nonumber\\
    &\propto \mathbb{P}[z_p|\omega_p=j,\mathbf{q}_{j},\mathbf{z}_{\nu(p)}] \mathbb{P}[\omega_p=j|\boldsymbol{\omega}_{\mathcal{V}(p)},c_p,\eta_p] \nonumber\\
    &\propto
      \begin{cases}
        \frac{q_{z_p,j} \eta_p \exp \left( \beta_2 \sum\limits_{p^\prime \in \nu(p)} \delta(j,\omega_{p^\prime}) \right)}{ \sum\limits_{k^\prime=1}^{K} q_{k^\prime,j} \exp \left( \beta_1 \sum\limits_{p^\prime \in \nu(p)} \delta(k^\prime,z_{p^\prime}) \right)}  &\text{when}\ \omega_p = c_p \\
        \frac{(1-\eta_p) q_{z_p,j} \exp \left( \beta_2 \sum\limits_{p^\prime \in \nu(p)} \delta(j,\omega_{p^\prime}) \right)}{ (C-1) \sum\limits_{k^\prime=1}^{K} q_{k^\prime,j} \exp \left( \beta_1 \sum\limits_{p^\prime \in \nu(p)} \delta(k^\prime,z_{p^\prime}) \right)} &\text{otherwise}
      \end{cases}
  \end{align}
  Note that, as for the sampling of the columns $\mathbf{q}_j$ ($j \in \mathcal{J}$) of the interaction matrix $\mathbf{Q}$, this conditional probability is considerably simplified when $\beta_1=0$ (i.e., when no spatial regularization is imposed on the cluster labels) since, in this case, $\sum\limits_{k^\prime=1}^{K} q_{k^\prime,j} \exp \left( \beta_1 \sum\limits_{p^\prime \in \nu(p)} \delta(k^\prime,z_{p^\prime}) \right)=1$.

\section{Application to hyperspectral image analysis}
\label{sec:appli}

  The proposed general framework introduced in the previous sections has been instanced for a specific application, namely the analysis of hyperspectral images. Hyperspectral imaging for Earth observation has been receiving increasing attention over the last decades, in particular in signal/image processing literatures \cite{Camps2014,Manolakis2014,Ma2014guest}. This keen interest of the scientific community can be easily explained by the richness of the information provided by such images. Indeed, generalizing the conventional red/green/blue color imaging, hyperspectral imaging collects spatial measurements acquired in a large number of spectral bands. Each pixel is associated with a vector of measurements, referred to as \emph{spectrum}, which characterizes the macroscopic components present in this pixel. Classification and spectral unmixing are two well-admitted techniques to analyze hyperspectral images. As mentioned earlier, and similarly to numerous applicative contexts, classifying hyperspectral images consists in assigning a discrete label to each pixel measurement in agreement with a predefined semantic description of the image. Conversely, spectral unmixing proposes to retrieve some elementary components, called \emph{endmembers}, and their respective proportions, called \emph{abundance} in each pixel, associated with the spatial distribution of the endmembers in over the scene~\cite{Bioucas-Dias2012}. Per se, spectral unmixing can be cast as a blind source separation or a nonnegative matrix factorization (NMF) task \cite{Ma2014}. The particularity of spectral unmixing, also known as spectral mixture analysis in the microscopy literature \cite{Dobigeon2012ultra}, lies in the specific constraints applied to spectral unmixing. As for any NMF problem, the endmembers signatures as well as the proportions are nonnegative. Moreover, specifically, to reach a close description of the pixel measurements, the abundance coefficients, interpreted as concentrations of the different materials, should sum to one for each spatial position.

  Nevertheless, yet complementary, these two classes of methods have been considered jointly in a very limited number of works~\cite{Dopido2014,Villa2011}. The proposed hierarchical Bayesian model offers a great opportunity to design a unified framework where these two methods can be conducted jointly. Spectral unmixing is perfectly suitable to be envisaged as the low-level task of the model described in Section \ref{sec:prob}. The abundance vector provides a biophysical description of a pixel which can be seen as a vector of latent variables of the corresponding pixel. The classification step is more related to a semantic description of the pixel. The low-level and clustering tasks of general framework described respectively in Sections \ref{sec:low-task} and \ref{sec:cluster}, are specified in what follows, while the classification task is directly implemented as in Section \ref{sec:high-task}.

\subsection{Bayesian model}

\noindent \textbf{Low-level interpretation: }  According to the conventional linear mixing model (LMM), the pixel spectrum $\mathbf{y}_p$ ($p \in \mathcal{P}$) observed in $d$ spectral bands are approximated by linear mixtures of $R$ elementary signatures $\mathbf{m}_r$ ($r=1,\ldots,R$), i.e.,
  \begin{equation}
  \label{eq:LMM}
    \mathbf{y}_p = \sum_{r=1}^R a_{r,p} \mathbf{m}_r + \mathbf{e}_p
  \end{equation}
  where $\mathbf{a}_p = \left[a_{1,p},\ldots,a_{R,p}\right]^T$ denotes the vector of mixing coefficients (or abundances) associated with the $p$th pixel and $\mathbf{e}_p$ is an additive error assumed to be white and Gaussian, i.e., $\mathbf{e}_p| s^2 \sim \mathcal{N}(\boldsymbol{0}_d, s^2\mathbf{I}_d)$. When considering the $P$ pixels of the hyperspectral image, the LMM can be rewritten with its matrix form
  \begin{equation}
  \label{eq:LMMmatrix}
    \mathbf{Y} = \mathbf{M}\mathbf{A} + \mathbf{E}
  \end{equation}
  where  $\mathbf{M}=[\mathbf{m}_1,\ldots,\mathbf{m}_R]$, $\mathbf{A} = \left[\mathbf{a}_1,\ldots,\mathbf{a}_P\right]$ and $\mathbf{E} = \left[\mathbf{e}_1,\ldots,\mathbf{e}_P\right]$ are the matrices of the endmember signatures, abundance vectors and noise, respectively.  In this work, the endmember spectra are assumed to be a priori known or previously recovered from the hyperspectral images by using an endmember extraction algorithm \cite{Bioucas-Dias2012}. Under this assumption, the LMM matrix formulation defined by \eqref{eq:LMMmatrix} can be straightforwardly interpreted as a particular instance of the low-level interpretation \eqref{eq:model_lat} by choosing the latent function $f_{\mathrm{lat}}(\cdot)$ as a linear mapping $f_{\mathrm{lat}}(\mathbf{A}) = \mathbf{M} \mathbf{A}$ and the statistical model $\psi(\cdot,\cdot)$ as the Gaussian probability density function parametrized by the variance $s^2$.

  In this applicative example, since the error variance $s^2$ is a nuisance parameter and generally unknown, this hyperparameter is included within the Bayesian model and estimated jointly with the parameters of interest. More precisely, the variance $s^2$ is assigned a conjugate inverse-gamma prior and a non-informative Jeffreys hyperprior is chosen for the associate hyperparameter $\delta$
  \begin{equation}
    s^2|\delta \sim \mathcal{IG}(1,\delta), \quad \delta \propto \frac{1}{\delta} \mathbbm{1}_{\mathbb{R}^{+}}(\delta).
  \end{equation}
  These choices lead to the following inverse-gamma conditional posterior distribution
  \begin{equation}
    s^2|\mathbf{Y},\mathbf{A} \sim \mathcal{IG}\left(1+\frac{Pd}{2}, \frac{1}{2} \sum_{p=1}^{P} \|\mathbf{y}_p - \mathbf{M}\mathbf{a}_p\|^2\right)
  \end{equation}
which is easy to sample from, as an additional step within the Gibbs sampling scheme described in Section \ref{sec:mcmc}.\\

\noindent \textbf{Clustering: }  In the current problem, the latent modeling $\Phi(\cdot;\cdot)$ in \eqref{eq:prior_lat} is chosen as Gaussian distributions elected for the latent vectors $\mathbf{a}_p$ ($p\in \mathcal{P}$),
  \begin{equation}
    \mathbf{a}_p|z_p=k,\boldsymbol{\psi}_k,\boldsymbol{\Sigma}_k \sim \mathcal{N}(\boldsymbol{\psi}_k,\boldsymbol{\Sigma}_k)
  \end{equation}
  where $\boldsymbol{\psi}_k$ and $\boldsymbol{\Sigma}_k$  are the mean vector and covariance matrix associated with the $k$th cluster. This Gaussian assumption is equivalent of considering each high-level class as a mixture of Gaussian distributions in the abundance space. The covariance matrices are chosen as $\boldsymbol{\Sigma}_k = \text{diag}(\sigma_{k,1}^2,\dots,\sigma_{k,R}^2)$ where $\sigma_{k,1}^2,\dots,\sigma_{k,R}^2$ are a set of $R$ unknown hyperparameters. The conditional posterior distribution of the abundance vectors $\mathbf{a}_p$ can be finally expressed as follows
  \begin{align}
   p(\mathbf{a}_p&|z_p=k, \boldsymbol{y}_p, \boldsymbol{\psi}_k, \boldsymbol{\Sigma}_k) \propto \nonumber\\
   &|\boldsymbol{\Lambda}_k|^{-\frac{1}{2}} \exp \left( -\frac{1}{2} (\mathbf{a}_p - \boldsymbol{\mu}_k)^t \boldsymbol{\Lambda}_k^{-1} (\mathbf{a}_p - \boldsymbol{\mu}_k) \right)
  \end{align}
  where $\boldsymbol{\mu}_k = \boldsymbol{\Lambda}_k (\frac{1}{s^2} \mathbf{M}^t \mathbf{y}_p + \boldsymbol{\Sigma}_k^{-1}\boldsymbol{\psi}_k)$ and $\boldsymbol{\Lambda}_k = (\frac{1}{s^2} \mathbf{M}^t\mathbf{M} + \boldsymbol{\Sigma}_k^{-1})^{-1}$. It shows that the latent vector $\mathbf{a}_p$ associated with a pixel belonging to the $k$th cluster is distributed according to the multivariate Gaussian distribution $\mathcal{N}(\boldsymbol{\mu}_k,\boldsymbol{\Lambda}_k)$.

 Moreover the variances $\sigma_{k,r}^2$ are included into the Bayesian model by choosing conjugate inverse-gamma prior distributions
  \begin{equation}
    \sigma_{k,r}^2 \sim \mathcal{IG}(\xi,\gamma)
  \end{equation}
  where parameters $\xi$ and $\gamma$ have been selected to obtain vague priors ($\xi=1$,$\gamma=0.1$). It leads to the following conditional inverse-gamma posterior distribution
    \begin{equation}
      \sigma_{r,k}|\mathbf{A},\mathbf{z},\psi_{r,k} \sim \mathcal{IG}\left(\frac{n_k}{2} + \xi, \gamma + \sum_{p\in \mathcal{I}_k} \frac{(a_{r,k} - \psi_{r,k})^2}{2} \right)
    \end{equation}
  where $n_k$ is the number of samples in cluster $k$, and $\mathcal{I}_k \subset \mathcal{P}$ is the set of indexes of pixels belonging to the $k$th cluster (i.e., such that $z_p=k$).

  Finally, the prior distribution of the cluster mean $\boldsymbol{\psi}_k$ ($k \in \mathcal{K}$) is chosen as a Dirichlet distribution $\text{Dir}(\mathbf{1})$. Such a prior induces \emph{soft} non-negativity and sum-to-one constraints on $\mathbf{a}_p$. Indeed, these two constraints are generally admitted to describe the abundance coefficients since they represent proportions/concentrations. In this work, this constraint is not directly imposed on the abundance vectors but rather on their mean vectors, since $\mathrm{E}[\mathbf{a}_p|z_p=k] = \boldsymbol{\psi}_k$. The resulting conditional posterior distribution of the mean vector $\boldsymbol{\psi_k}$ is the following multivariate Gaussian distribution
  \begin{equation}
    \boldsymbol{\psi}_k|\mathbf{A},\mathbf{z}, \boldsymbol{\Sigma}_k \sim \mathcal{N}_{\mathbb{S}}\left(\frac{1}{n_k} \sum_{p\in \mathcal{I}_k} \mathbf{a}_p, \frac{1}{n_k} \boldsymbol{\Sigma}_k\right)
  \end{equation}
  truncated on the probability simplex
  \begin{equation}
    \mathbb{S} = \left\{\mathbf{x}=[x_1,\ldots,x_R]^T | \forall r,\ x_r \geq 0 \ \text{and} \ \sum_{r=1}^R x_r =1\right\}.
  \end{equation}
 Sampling according to this truncated Gaussian distribution can be achieved following the strategies described in \cite{Altmann2014ssp}.


\subsection{Experiments}

\subsubsection{Synthetic dataset}
\label{sec:synthetic}
\begin{figure}[!ht]
    \centering
    \begin{tabular}{@{}cc}
      \includegraphics[width=0.45\columnwidth]{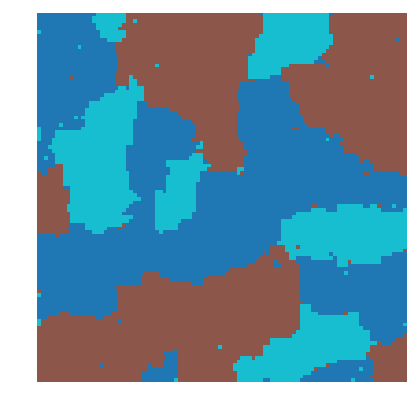}&
      \includegraphics[width=0.45\columnwidth]{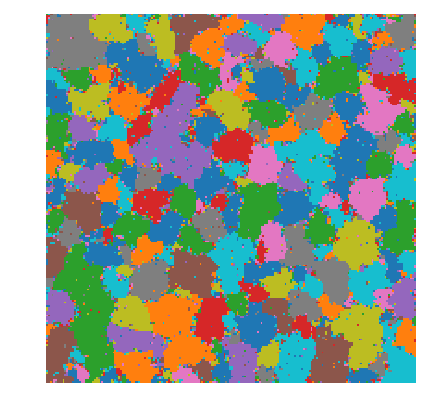}\\
      (a) & (b) \\
      \includegraphics[width=0.45\columnwidth]{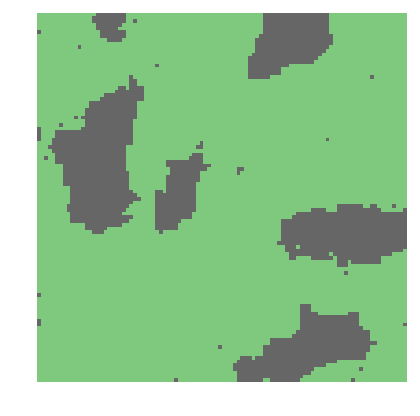}&
      \includegraphics[width=0.45\columnwidth]{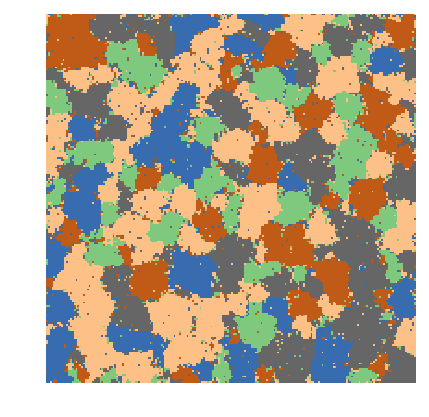}\\
      (c) & (d) \\
    \end{tabular}
    \caption{Synthetic data. Classification maps of Image 1 (a) and Image 2 (b), corresponding clustering maps of Image 1 (c) and Image 2 (d).\label{fig:synth-data}}
  \end{figure}

Synthetic data have been used to assess the performance of the proposed analysis model and algorithm. Hyperspectral images have been synthetically generated according to the following hierarchical procedure. First, cluster maps have been generated from Potts-Markov MRFs. Then, the corresponding classification maps have then been chosen by artificially merging a few of these clusters to define each class. Abundance vectors in a given cluster have been randomly drawn from a Dirichlet distribution parametrized by a specific mean for each cluster. Finally the pixel measurements have been generated using the linear mixture model with real endmembers signatures of $d=413$ spectral bands extracted from a spectral library. These linearly mixed pixels have been corrupted by a Gaussian noise resulting in a signal-to-noise ratio of SNR$=30$dB. Two distinct images, referred to as Image 1 and Image 2 and represented in Figure~\ref{fig:synth-data}, have been considered. The first one is a $100\times100$-pixel image composed of $R=3$ endmembers, $K=3$ clusters and $J=2$ classes. The second hyperspectral image is a $200\times200$-pixel image which consists of $R=9$ endmembers, $K=12$ clusters and $J=5$ classes.

\begin{figure}[!ht]
    \centering
    \begin{tabular}{@{}cc}
      \includegraphics[width=0.45\columnwidth]{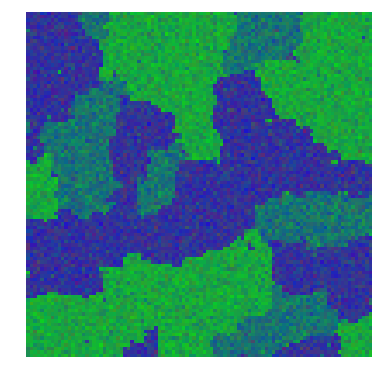}&
      \includegraphics[width=0.45\columnwidth,height=0.45\columnwidth]{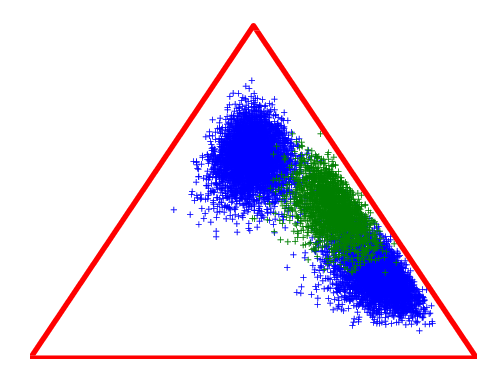}\\
    \end{tabular}
    \caption{Image 1. Left: colored composition of abundance map. Right: pixels in the probabilistic simplex (red triangle) with Class 1 (blue) and Class 2 (green).\label{fig:abund-simplex}}
  \end{figure}

  Figure~\ref{fig:abund-simplex} represents the abundance vectors of each pixel in the probabilistic simplex for Image 1. The three clusters are clearly identifiable and the class represented in blue is also clearly divided into two clusters.

  To evaluate the interest of including the classification step into the model, results provided by the proposed method have been compared to the counterpart model proposed in \cite{Eches2011} (referred to as Eches model) which does not exploit this high-level information. The pixels and associated classification labels located in the upper quarters of the Images 1 and 2 have been used as the training set $\mathcal{L}$. The confidence in this classification ground-truth has been set to a value of $\eta_p=0.95$ for all the pixels ($p\in \mathcal{L}$). Additionally, the values of Potts-MRF granularity parameters have been selected as $\beta_1=\beta_2=0.8$. In the case of the Eches model, the images have been subsequently classified using the estimated abundance vectors and clustering maps, and following the strategy proposed in~\cite{Bouveyron2009}. The performance of the spectral unmixing task has been evaluated using the root global mean square error (RGMSE) associated with the abundance estimation
  \begin{equation}
    \mathrm{RGMSE}(\mathbf{A}) = \sqrt{\frac{1}{PR} \left\|\hat{\mathbf{A}}-\mathbf{A}\right\|_F^2}
  \end{equation}
  where $\hat{\mathbf{A}}$ and $\mathbf{A}$ denote respectively the estimated and actual matrices of abundance vectors. Moreover, the accuracy of the estimated classification maps has been measured with the conventional Cohen's kappa. Results reported in Table~\ref{tab:res-metric} show that the obtained RMSE are not significantly different between the two models. Moreover, the comparison between processing times shows a small computational overload required by the proposed model. It should be noticed that this experiment has been conducted with a fixed number of iterations of the proposed MCMC algorithm ($300$ iterations including $50$ burn-in iterations).

  \begin{table*}
    \centering
    \caption{Unmixing and classification results for various datasets.\label{tab:res-metric}}
    \begin{tabular}[b]{llccccc}\toprule
          & & RGMSE$(\mathbf{A})$  & Kappa & Time (s) \\
          \midrule
          \multirow{2}{*}{Image 1} & Proposed model & 3.23e-03 (1.6e-05) & 0.932 (0.018) & 171 (5.4) \\
                                   & Eches model    & 3.24e-03 (1.4e-05) & 0.909 (0.012) & 146 (0.7) \\
          \midrule
          \multirow{2}{*}{Image 2} & Proposed model & 1.62e-02 (1.62e-04) & 0.961 (0.04)   & 950 (11)  \\
                                   & Eches model    & 1.61e-02 (2.71e-05) & 0.995 (0.0004) & 676 (2.1)  \\
          \midrule
          MUESLI image             & Proposed model & N$\backslash$ A     & 0.856 (0.004)  & 5472 (84) \\
          \bottomrule
    \end{tabular}
  \end{table*}

\begin{figure}[!ht]
    \centering
      \begin{tikzpicture}
        \begin{axis}[width=0.9\columnwidth,ymin=0.,ymax=1.,xmin=0.,grid,axis x line=left,axis y line=left,xlabel={$\alpha$ (label corruption in \%)},ylabel={kappa},ylabel style={align=center},font=\footnotesize]
          \addplot[thick,red] table[x expr=1-\thisrowno{6},y index=0] {conv_labelRatio_K3_beta0.8.txt};
          \addplot[name path=A, draw=none] table[x expr=1-\thisrowno{6},y expr=\thisrowno{0}+\thisrowno{3}] {conv_labelRatio_K3_beta0.8.txt};
          \addplot[name path=B, draw=none] table[x expr=1-\thisrowno{6},y expr=\thisrowno{0}-\thisrowno{3}] {conv_labelRatio_K3_beta0.8.txt};
          \addplot[red!80, opacity=0.3] fill between[of=A and B];
          \addplot[thick,blue] table[x expr=1-\thisrowno{6},y index=1] {conv_labelRatio_K3_beta0.8.txt};
          \addplot[name path=C, draw=none] table[x expr=1-\thisrowno{6},y expr=\thisrowno{1}+\thisrowno{4}] {conv_labelRatio_K3_beta0.8.txt};
          \addplot[name path=D, draw=none] table[x expr=1-\thisrowno{6},y expr=\thisrowno{1}-\thisrowno{4}] {conv_labelRatio_K3_beta0.8.txt};
          \addplot[blue!80, opacity=0.3] fill between[of=C and D];
          \addplot[thick,green] table[x expr=1-\thisrowno{6},y index=2] {conv_labelRatio_K3_beta0.8.txt};
          \addplot[name path=E, draw=none] table[x expr=1-\thisrowno{6},y expr=\thisrowno{2}+\thisrowno{5}] {conv_labelRatio_K3_beta0.8.txt};
          \addplot[name path=F, draw=none] table[x expr=1-\thisrowno{6},y expr=\thisrowno{2}-\thisrowno{5}] {conv_labelRatio_K3_beta0.8.txt};
          \addplot[green!80, opacity=0.3] fill between[of=E and F];
        \end{axis};
      \end{tikzpicture}
    \caption{Classification accuracy measured with Cohen's kappa as a function of label corruption $\alpha$: proposed model (\textcolor{red}{red}), MDA with abundance vectors (\textcolor{blue}{blue}) and MDA with measured reflectance (\textcolor{green}{green}). Shaded areas denote the intervals corresponding to the standard deviation computed over $20$ trials.\label{fig:corrup-influence}}
  \end{figure}
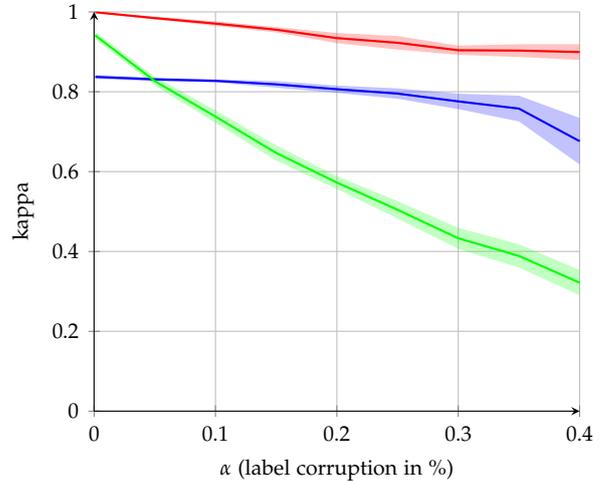

  A second scenario is considered where the training set includes label errors. The corrupted training set is generated by tuning a varying probability $\alpha$ to assign an incorrect label, all the other possible labels being equiprobable. The probability $\alpha$ varies from $0$ to $0.4$ with a $0.05$ step. In this context, the confidence in the classification ground-truth map is set equal to $\eta_p = 1-\alpha$ ($\forall p \in \mathcal{L}$). The results, averaged over $20$ trials for each setting, are compared to the results obtained using a mixture discriminant analysis (MDA)~\cite{Hastie1996} conducted either directly on the pixel spectra, either on the abundance vectors estimated with the proposed model. The resulting classification performances are depicted in Figure~\ref{fig:corrup-influence} as function of $\alpha$. These results show that the proposed model performs very well even when the training set is highly corrupted (i.e., $\alpha$ close to $0.4$).

\begin{figure}
    \centering
    \begin{tikzpicture}
        \begin{axis}[
            title={Image 1 estimated $\mathbf{Q}$},
            xlabel=Cluster,xtick={1,2,3},
            ylabel=Class,ytick={1,2},
            colorbar,
            colorbar style={
            },
            width=5cm,height=3.5cm,
            xmin=0.5,xmax=3.5,ymin=0.5,ymax=2.5,
            colormap/viridis,
            %
            enlargelimits=false,
        ]
          \addplot [matrix plot*,point meta=explicit] file {Q_K3_expl.txt};
        \end{axis}
    \end{tikzpicture}
    \begin{tikzpicture}
        \begin{axis}[
            title={Image 2 estimated $\mathbf{Q}$},
            xlabel=Cluster,xtick={1,3,5,7,9,11},
            ylabel=Class,ytick={1,3,5},
            colorbar,
            colorbar style={
            },
            width=5cm,height=3.5cm,
            xmin=0.5,xmax=12.5,ymin=0.5,ymax=5.5,
            colormap/viridis,
            %
            enlargelimits=false,
        ]
          \addplot [matrix plot*,point meta=explicit] file{Q_K12_expl.txt};
        \end{axis}
    \end{tikzpicture}
    \caption{Estimated interaction matrix $\mathbf{Q}$ for Image 1 (top) and Image 2 (bottom).\label{fig:q-expl}}
  \end{figure}
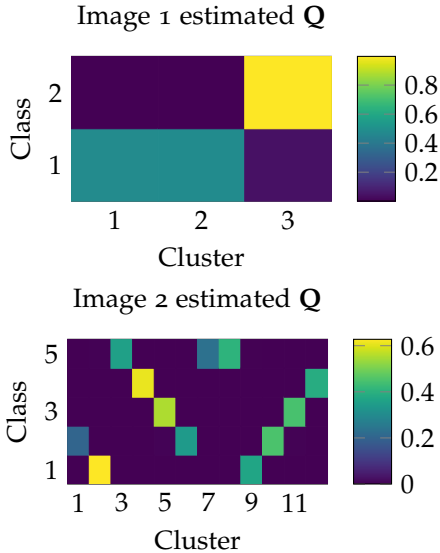

  Moreover, as already explained, another advantage of the proposed model is the interesting by-products provided by the method. As an illustration, Figure~\ref{fig:q-expl} presents the interactions matrices $\mathbf{Q}$ estimated for each image. From this figure, it is clearly possible to identify the structure of the various classes and their hierarchical relationship with the underlying clusters. For instance, for Image 2, it can be noticed that Class $1$ is essentially composed of the clusters $\sharp 2$ and $\sharp 9$.

\subsubsection{Real hyperspectral image}
\begin{figure}
    \centering
    \begin{tabular}{@{}cc}
      \includegraphics[width=0.45\columnwidth]{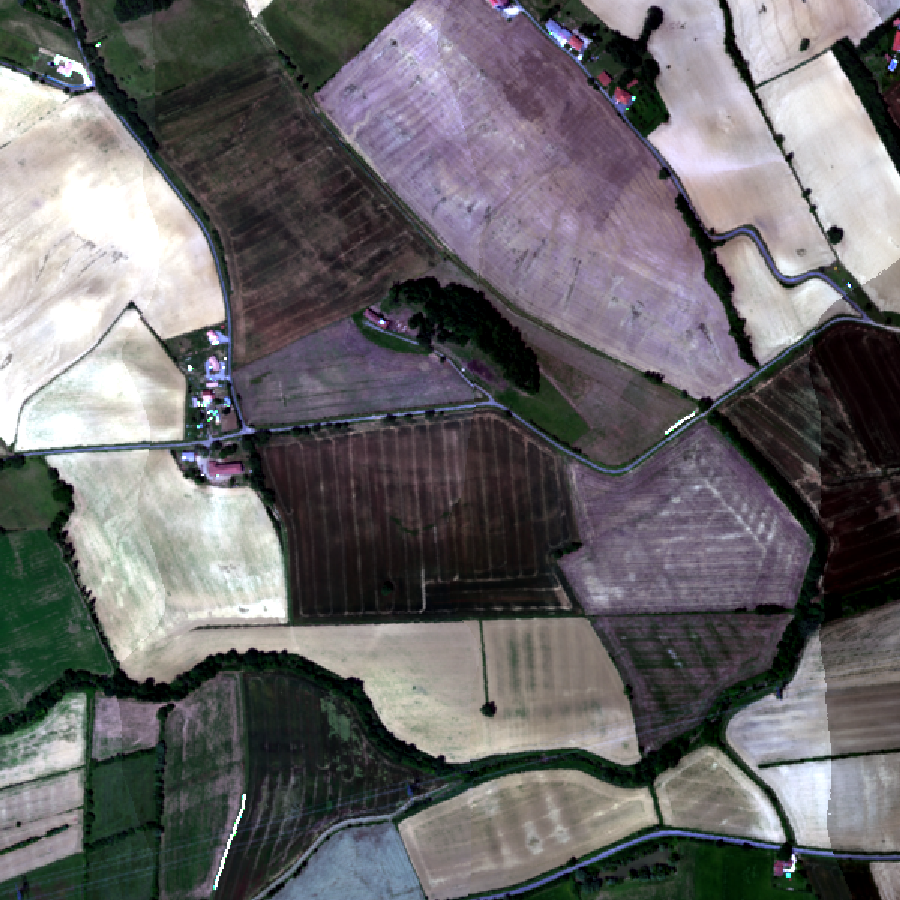}&
      \includegraphics[width=0.45\columnwidth]{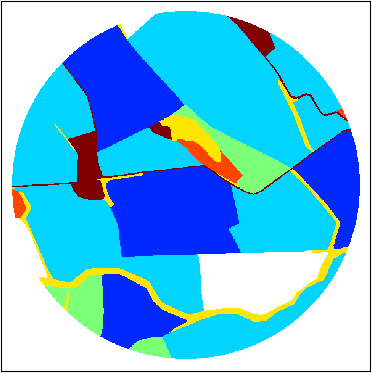}\\
      (a) & (b) \\
      \includegraphics[width=0.45\columnwidth]{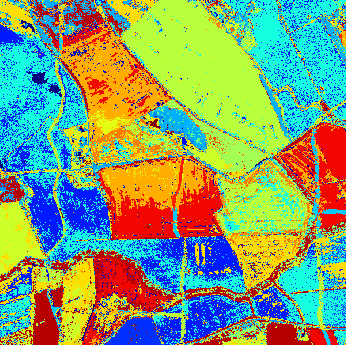}&
      \includegraphics[width=0.45\columnwidth]{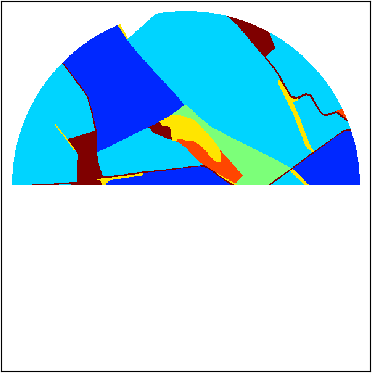}\\
      (c) & (d) \\
      \includegraphics[width=0.45\columnwidth]{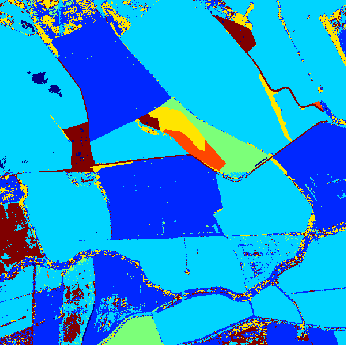}&
      \includegraphics[width=0.45\columnwidth]{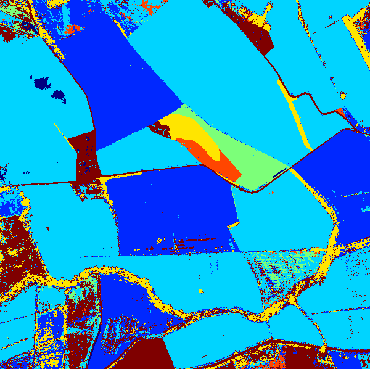}\\
      (e) & (f) \\
    \end{tabular}
    \caption{Real MUESLI image. Colored composition of the hyperspectral image (a), expert ground-truth (b), estimated clustering (c), training data (d), estimated classification with proposed model (e) and estimated classification with random forest (f).\label{fig:real-data}}
  \end{figure}

  Finally, the proposed strategy has been implemented to analyze a real $600\times600$-pixel hyperspectral image acquired within the framework of the \emph{multiscale mapping of ecosystem services by very high spatial resolution hyperspectral and LiDAR remote sensing imagery} (MUESLI) project \footnote{\url{http://fauvel.mathieu.free.fr/pages/muesli.html}}. This image is composed of $d=438$ spectral bands and $R=7$ endmembers have been extracted using the widely-used vertex component analysis (VCA) algorithm~\cite{Nascimento2005}. The associated expert ground-truth classification is made of $6$ classes (straw cereals, summer crops, wooded area, buildings, bare soil, pasture). In this experiment, the upper half of the expert ground-truth has been provided as training data for the proposed method. The confidence $\eta_p$ has been set to $95$\% for all training pixels to account for the imprecision of the expert ground-truth. The MRF granularity parameters of the proposed parameters have been set to $\beta_1=0.3$ and $\beta_2=1$ since these values provide the most meaningful interpretation of the image. Figure~\ref{fig:real-data} presents a colored composition of the hyperspectral image (a), the expert ground-truth (b) and the obtained results in terms of clustering (c) and classification (d). Quantitative results in term of classification accuracy have been computed and are summarized in Table~\ref{tab:res-metric}. Note that no performance measure of the unmixing step is provided since no abundance groundtruh is available for this real dataset.
  
\begin{figure}[!ht]
    \centering
      \begin{tikzpicture}
        \begin{axis}[width=0.9\columnwidth,ymin=0.,ymax=1.,grid,axis x line=left,axis y line=left,xlabel={$\alpha$ (label corruption in \%)},ylabel={kappa},ylabel style={align=center},font=\footnotesize]
          \addplot[thick,red] table[x index=4,y index=7] {table_mod2_K40_fus0.0_labprop0.5_equiprobC0_R7.txt};
          \addplot[name path=A, draw=none] table[x index=4,y expr=\thisrowno{7}+\thisrowno{11}] {table_mod2_K40_fus0.0_labprop0.5_equiprobC0_R7.txt};
          \addplot[name path=B, draw=none] table[x index=4,y expr=\thisrowno{7}-\thisrowno{11}] {table_mod2_K40_fus0.0_labprop0.5_equiprobC0_R7.txt};
          \addplot[red!80, opacity=0.3] fill between[of=A and B];
          \addplot[thick,blue] table[x index=1,y index=4] {table_rf.txt};
          \addplot[name path=C, draw=none] table[x index=1,y expr=\thisrowno{4}+\thisrowno{8}] {table_rf.txt};
          \addplot[name path=D, draw=none] table[x index=1,y expr=\thisrowno{4}-\thisrowno{8}] {table_rf.txt};
          \addplot[blue!80, opacity=0.3] fill between[of=C and D];
        \end{axis};
      \end{tikzpicture}
    \caption{Real MUESLI image. Classification accuracy measured with Cohen's kappa as a function of label corruption $\alpha$: proposed model (\textcolor{red}{red}), random forest (\textcolor{blue}{blue}). Shaded areas denote the intervals corresponding to the standard deviation computed over $10$ trials.\label{fig:corrup-influence-muesli}}
  \end{figure}
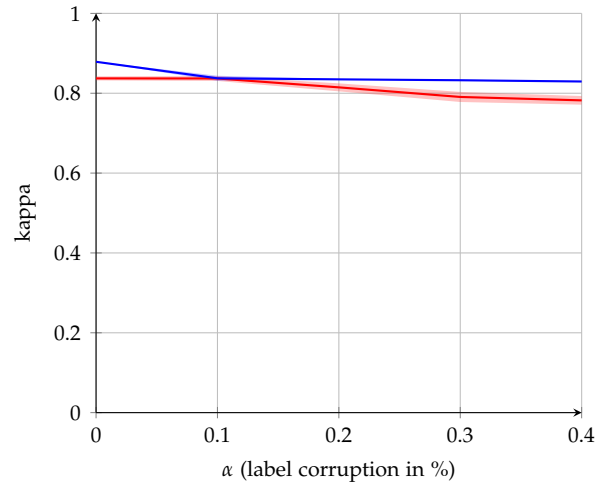

  Additionally, the robustness with respect to expert mislabeling of the ground-truth training dataset has been evaluated and compared to the performance obtained by a state-of-the-art random forest (RF) classifier. Errors in the expert ground-truth have been randomly generated with the same process as the one used for the previous experiment with synthetic data (see Section \ref{sec:synthetic}). Confidence in the ground-truth has been set equal to $\eta_p=1-\alpha$ for all the pixels ($p\in \mathcal{L}$) where $\alpha$ is the corruption rate, with a maximum of $95$\% of confidence. Parameters of the RF classifier have been optimized using cross-validation on the training set. Classification accuracy measured through Cohen's kappa is presented in Figure~\ref{fig:corrup-influence-muesli} as a function of the corruption rate $\alpha$ of the training set. From these results, the proposed method seems to perform favorably when compared to the RF classifier. It is worth noting that RF is one of the prominent method to classify remote sensing data and that the robustness to noise in labeled data is a well-documented property of this classification technique~\cite{Pelletier2017}.

\section{Conclusion and perspectives}
\label{sec:ccl}

This paper proposed a Bayesian model to perform jointly low-level modeling and robust classification. This hierarchical model capitalized on two Markov random fields to promote coherence between the various levels defining the model, namely, i) between the clustering conducted on the latent variables of the low-level modeling and the estimated class labels, and ii) between the estimated class labels and the expert partial label map provided for supervised classification. The proposed model was specifically designed to result into a classification step robust to labeling errors that could be present in the expert ground-truth. Simultaneously, it offered the opportunity to correct mislabeling errors. This model was particularly instanced on a particular application which aims at conducting hyperspectral image unmixing and classification jointly. Numerical experiments were conducted first on synthetic data and then on real data. These results demonstrate the relevance and accuracy of the proposed method. The richness of the resulting image interpretation was also underlined by the results. Future works include the generalization of the proposed model to handle fully unsupervised low-level analysis tasks. Instantiations of the proposed model in other applicative contexts will be also considered.

\section*{References}

\bibliographystyle{elsarticle-num}
\bibliography{strings_all_ref,biblio}

\end{document}